\documentclass[5p]{elsarticle} 

\usepackage{amssymb}
\usepackage{float}
\usepackage{array}
\usepackage{booktabs}
\usepackage[table]{xcolor}
\usepackage{bbding}
\usepackage{booktabs}

\usepackage{lineno}
\usepackage{multirow}
\usepackage{amsmath,amssymb,amsfonts}
\usepackage{algorithmic}
\usepackage{graphicx}
\usepackage{textcomp}
\usepackage{color}
\journal{Information Fusion}

\usepackage{ecrc}
\volume{00}
\firstpage{1}
\journalname{Information Fusion}
\runauth{}
\jid{procs}
\jnltitlelogo{Information Fusion}
\CopyrightLine{2021}{Published by Elsevier Ltd.}

\newcommand{\colorbibs}[2][blue]%
{%
	\DeclareBibliographyCategory{ColoredBiblist#1}%
	\addtocategory{ColoredBiblist#1}{#2}%
	\AtEveryBibitem{\ifcategory{ColoredBiblist#1}{\color{#1}\bfseries}{}}
}

\begin{document}
\begin{sloppypar}

\begin{frontmatter}



\title{CILF-CIAE: CLIP-driven Image–Language Fusion for Correcting  Inverse Age Estimation}


\author[address1]{Yuntao Shou}
\ead{shouyuntao@stu.xjtu.edu.cn}

\author[address1]{Tao Meng\corref{cor1}}
\ead{mengtao@hnu.edu.cn}
\cortext[cor1]{Corresponding author}

\author[address1]{Wei Ai}
\ead{aiwei@hnu.edu.cn}


\author[address3]{Nan Yin}
\ead{nan.yin@mbzuai.ac.ae}

\author[address1]{Fuchen Zhang}
\ead{fuchen.zhang@csuft.edu.cn}

\author[address2]{Keqin Li}
\ead{lik@newpaltz.edu}

\address[address1]{College of Computer and Information Engineering, Central South University of Forestry and Technology, Hunan, China}

\address[address2]{Department of Computer Science, State University of New York, New Paltz, New York 12561, USA}

\address[address3]{Mohamed bin Zayed University of Artificial Intelligence,
UAE}

\begin{abstract}
The age estimation task aims to predict the age of an individual by analyzing facial features in an image. The development of age estimation can improve the efficiency and accuracy of various applications (e.g., age verification and secure access control, etc.). In recent years, contrastive language-image pre-training (CLIP) has been widely used in various multimodal tasks and has made some progress in the field of age estimation. However, existing CLIP-based age estimation methods require high memory usage (quadratic complexity) when globally modeling images, and lack an error feedback mechanism to prompt the model about the quality of age prediction results. To tackle the above issues, we propose a novel CLIP-driven Image–Language Fusion for Correcting Inverse Age Estimation (CILF-CIAE). Specifically, we first introduce the CLIP model to extract image features and text semantic information respectively, and map them into a highly semantically aligned high-dimensional feature space. Next, we designed a new Transformer architecture (i.e., FourierFormer) to achieve channel evolution and spatial interaction of images, and to fuse image and text semantic information. Compared with the quadratic complexity of the attention mechanism, the proposed Fourierformer is of linear log complexity. To further narrow the semantic gap between image and text features, we utilize an efficient contrastive multimodal learning module that supervises the multimodal fusion process of FourierFormer through contrastive loss for image-text matching, thereby improving the interaction effect between different modalities. Finally, we introduce reversible age estimation, which uses end-to-end error feedback to reduce the error rate of age predictions. Through extensive experiments on multiple data sets, CILF-CIAE has achieved better age prediction results.
\end{abstract}


\end{frontmatter}

\section{Introduction}
\subsection{Motivation}
The task of age estimation aims to determine the age based on the facial features in the image. In recent years, due to the massive increase in image data sets and the widespread application of deep learning (DL), age estimation methods have also achieved important achievements and attracted widespread research attention \cite{shen2019deep}, \cite{liu2020similarity}, \cite{10113198}. Futhermore, age estimation is also widely used in many scenarios. For example, age estimation in finance and insurance can help detect fraud where age is falsely stated to obtain improper benefits \cite{rothe2018deep}, \cite{bao2023deep}, \cite{10314020}.

\begin{figure}
	\centering
	\includegraphics[width=1\linewidth]{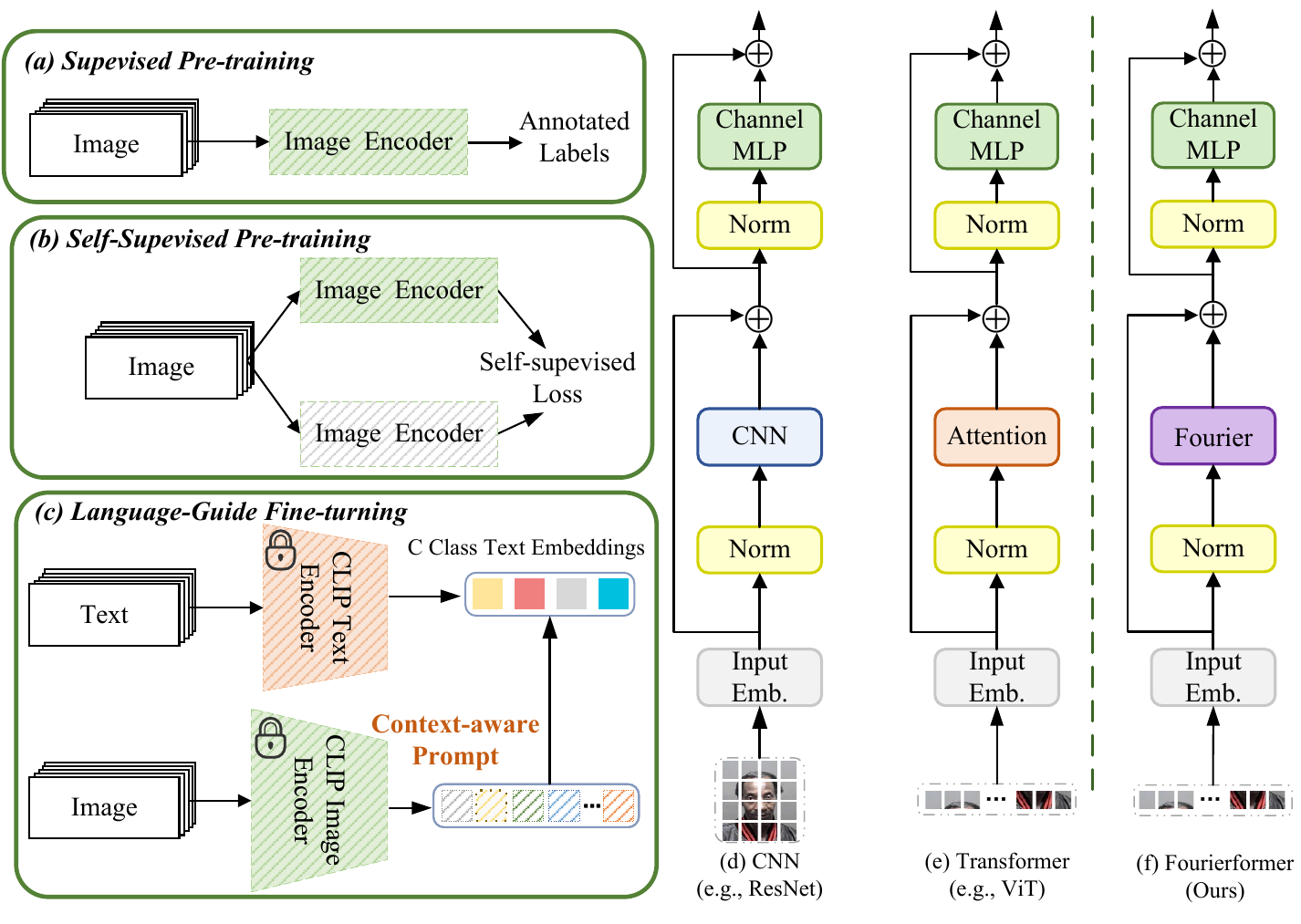}
	\caption{We compare the differences between existing image processing paradigms and the paradigm proposed in this paper. As shown in Fig. \ref{fig:paradigm}(a), most image processing methods perform supervised learning by inputting images and then using manually annotated labels as supervision signals. As shown in Fig. 1(b), since manual annotation requires a large amount of resources, existing methods begin to build self-supervised learning models by contrasting input images. As shown in Fig. \ref{fig:paradigm}(c), we perform text-image contrastive learning by using the CLIP pre-trained model and transfer the learned knowledge to the age estimation prediction task. As shown in Figs \ref{fig:paradigm} (d) and (e), existing methods are mainly based on CNN architecture and Transformer architecture based on attention mechanism to extract feature information of images. As shown in Fig. 1(f), we replace the attention module in the Transformer architecture with a Fourier prior module.}
	\label{fig:paradigm}
\end{figure}

The current mainstream age estimation methods are divided into three categories: CNN \cite{niu2016ordinal}, \cite{duan2017ensemble}, attention network \cite{wang2022improving}, \cite{zhang2019fine}, and GCN \cite{shou2023masked}. To extract global information and multi-scale information in images, a CNN-based age estimation algorithm is applied. For example, Rothe et al. \cite{rothe2018deep} estimated an individual's true age and apparent age from a single face image based on a CNN method. Unlike many traditional machine learning methods \cite{cao2012human}, this method does not require the use of facial feature point markers and only requires the input of face images for age estimation. However, CNN-based methods cannot capture the semantic features in images that are most relevant to age features. To give higher weight to the semantic features in the image that are most relevant to the age feature, attention networks began to be applied. For instance, Shen et al. \cite{shen2022attention} introduced an attention mechanism so that the model can automatically focus on regions in the image that are relevant for age estimation, which helps improve the model's perception of important features related to age. However, attention network-based methods cannot flexibly model irregular objects. To overcome the above problems, Shou et al. \cite{shou2023masked} proposed a contrastive multi-view GCN for age estimation (CMGCN). CMGCN improves the feature representation capabilities of images by extending image representation into topological semantic space. However, the methods mentioned above are all supervised learning methods and ignore the CLIP-based multimodal learning paradigm. Taking Fig \ref{fig:paradigm}(a) and (b) as an example, existing age estimation algorithms mainly focus on supervised, or self-supervised algorithm design \cite{bao2022divergence}, \cite{deng2021pml}, ignoring the contrastive image-language pre-training (CLIP) paradigm. CLIP can learn the prior information of faces from a large number of text-image pairs and provide better generalization for downstream tasks. Specifically, CLIP learns the correlation between images and text from a large number of image-text pairs through contrastive learning. Furthermore, existing algorithms directly predict age and lack an error information feedback mechanism, which may lead to a large error between the model's predicted age and the true label. Therefore, it is necessary to take CLIP multimodal learning paradigm and error-controllable generation as the starting point for model design.

To tackle the above problem, we propose a novel CLIP-driven Image–Language Fusion for Correcting Inverse Age Estimation (CILF-CIAE) to perform age estimation. CILF-CIAE mainly includes four modules: CLIP-based visual and language feature encoder, Fourierformer-based feature fusion, age prediction and error-controllable generation module. Firstly, we use Image Encoder and Text Encoder in CLIP to encode image and text features respectively and obtain corresponding feature representations. After obtaining the image and text feature representations, we jointly input them into the $N$-dimensional feature space for contrastive learning to obtain aligned text and image semantic vectors, and utilize obtained image semantic vectors to perform age estimation. Secondly, as shown in Fig. \ref{fig:paradigm}(d) and (e), unlike previous CNN-based and attention-based Transformer architectures, CNN-based methods can only extract local information of the image and it is difficult to use contextual prompts modules to enhance age estimation, while attention-based methods require large memory usage (quadratic complexity). We introduce the Transformer architecture based on Fourier transform to realize the spatial interaction and channel evolution of image features, so as to fuse text and image feature information to improve the age estimation performance. Specifically, we replace the attention module in Transformer with Fourier transform and input image features into Fuorierformer to achieve spatial interaction and channel evolution. To further narrow the semantic gap between image and text features, we utilize an efficient contrastive multimodal learning module that supervises the multimodal fusion process of FourierFormer through contrastive loss for image-text matching, thereby improving the interaction effect between different modalities. Thirdly, we construct age estimation prediction loss and text and image matching loss to complete the parameter optimization of the model. Finally, we build an error-correcting reversible age estimation module to ensure that the predicted age is within a high-confidence interval in an end-to-end learning manner.

\subsection{Our Contributions}
Therefore, CLIP multimodal learning, spatial interaction of images, and channel evolution should be the core of age estimation algorithm design. Inspired by the above analysis, we propose a novel CLIP-driven Image–Language Fusion for Correcting Inverse Age Estimation (CILF-CIAE) to perform age estimation. The main contributions of this paper are summarized as follows:

\begin{enumerate}
	\item A novel CLIP-driven Image–Language Fusion for Correcting Inverse Age Estimation architecture is present and named CILF-CIAE. CILF-CIAE is able to learn information about age from input images.
	
	\item A new Transformer structure is designed, i.e., Fourierformer. FourierFormer replaces the attention mechanism with Fourier transform to realize the channel evolution and spatial interaction of image features.
	
	\item An efficient contrastive multimodal learning module is utilized to supervise the multimodal fusion process of FourierFormer through contrastive loss for image-text matching, thereby improving the interaction effect between different modalities.
	
	\item An end-to-end error feedback mechanism is proposed to ensure that the confidence of age estimation is within a credible range.
	
	\item Extensive experiments are conducted on four real data sets to verify the effectiveness of the method CILF-CIAE proposed in this paper. Experimental results show that CILF-CIAE can achieve optimal age prediction.
\end{enumerate}

\section{Related work}

\subsection{Age Estimation}
Traditional age estimation methods usually rely on hand-designed feature extraction and machine learning algorithms, which are limited by feature selection and age estimation performance \cite{cao2012human}, \cite{yin2023coco}, \cite{10.1145/3503161.3548012}. With the popularity of the Internet and social media (e.g., meta, twitter, and Youtube, etc.), large-scale face image datasets have also been widely grown. The rapid growth of data sets provides rich training data for deep learning (DL), making DL's learning capabilities more powerful. Age estimation has potential applications in social media analysis, ad targeting, security monitoring, medical image analysis, etc. For example, in security and legal applications, image age estimation can assist police in identifying possible underage criminal suspects.

Existing age estimation algorithms are mainly divided into two categories, i.e., age estimation algorithms based on machine learning and algorithms based on deep learning. Machine learning-based age estimation algorithms mainly rely on hand-designed rules to extract age-related features of images. Age estimation algorithms based on deep learning mainly use some deep learning models (e.g., CNN, Transformer, and GCN, etc) with powerful adaptive learning capabilities and massive data sets to estimate age in an end-to-end manner.

\textbf{Machine learning methods:} In the age estimation algorithms based on traditional machine learning algorithms, Shin et al. \cite{shin2022moving} proposed an ordinal regression algorithm (MWR) based on moving window regression, which first ranks the input and reference labels and designs global and local regressors to achieve prediction of global ranking and local ranking. MWR achieves fine-grained age estimation by continuously iteratively optimizing the ranking order. However, the computational complexity of MWR is relatively high. Cao et al. \cite{cao2020rank} proposed a consistent ranking logic algorithm to solve the inconsistency problem of multiple binary ordinal regression algorithms. CORAL ensures ranking consistency by introducing confidence scores. Cao et al. \cite{cao2012human} proposed the Ranking SVM algorithm to achieve age estimation of images. This algorithm estimates age by first grouping ages and then sorting ages. RSVM can reduce the hypothesis space of model learning. Zhang et al. \cite{zhang2017quantifying} achieved age estimation by learning the probability distribution of label information. This algorithm achieves age prediction by calculating the posterior probability of the image. There are some other typical traditional machine learning algorithms \cite{li2019bridgenet}, \cite{shen2018deep}.

\textbf{Deep learning methods:} In the age estimation algorithms based on deep learning algorithms, CNN \cite{levi2015age}, attention network \cite{wang2022improving}, and hybrid neural network systems \cite{xie2015hybrid} are currently common age estimation algorithms. For example, Levi et al. \cite{levi2015age} proposed an age estimation algorithm based on deep CNN to solve the problem of insufficient performance of traditional machine learning algorithms. DeepCNN can achieve better prediction results even on a small amount of data sets. Duan et al. \cite{duan2017ensemble} proposed the CNN2ELM algorithm to combine the advantages of CNN and regression algorithms. CNN2ELM constructed three feature extraction networks of age, gender, and race, and used a fusion mechanism to fuse the complementary information of the three networks, and used ELM for regression prediction of age. Wang et al. \cite{wang2022improving} proposed the Attention-based Dynamic Patch Fusion algorithm to solve the problem that CNN cannot extract the most beneficial semantic information in the image for the age estimation task. ADPF introduces attention network and fusion network to dynamically extract image patches with rich semantic features and adaptively fuse the extracted feature information. Zhang et al. \cite{zhang2019fine} proposed a fine-grained attention LSTM algorithm to solve the problem that existing methods only focus on the global information of the image and ignore the fine-grained features of the image. This method first uses the residual network to extract the global information of the image, and then uses the attention LSTM to capture the sensitive area information of the image to obtain local important semantic features in the image. Xie et al. \cite{xie2015hybrid} integrated CNN's feature extraction capabilities, domain generalization capabilities, and local information discrimination capabilities based on dictionary algorithms. This method first uses a pre-trained CNN to extract the feature representation of the image, and then builds a dictionary representation to extract the local feature information and Fisher vector representation of the image.

\begin{figure*}
	\centering
	\includegraphics[width=1\linewidth]{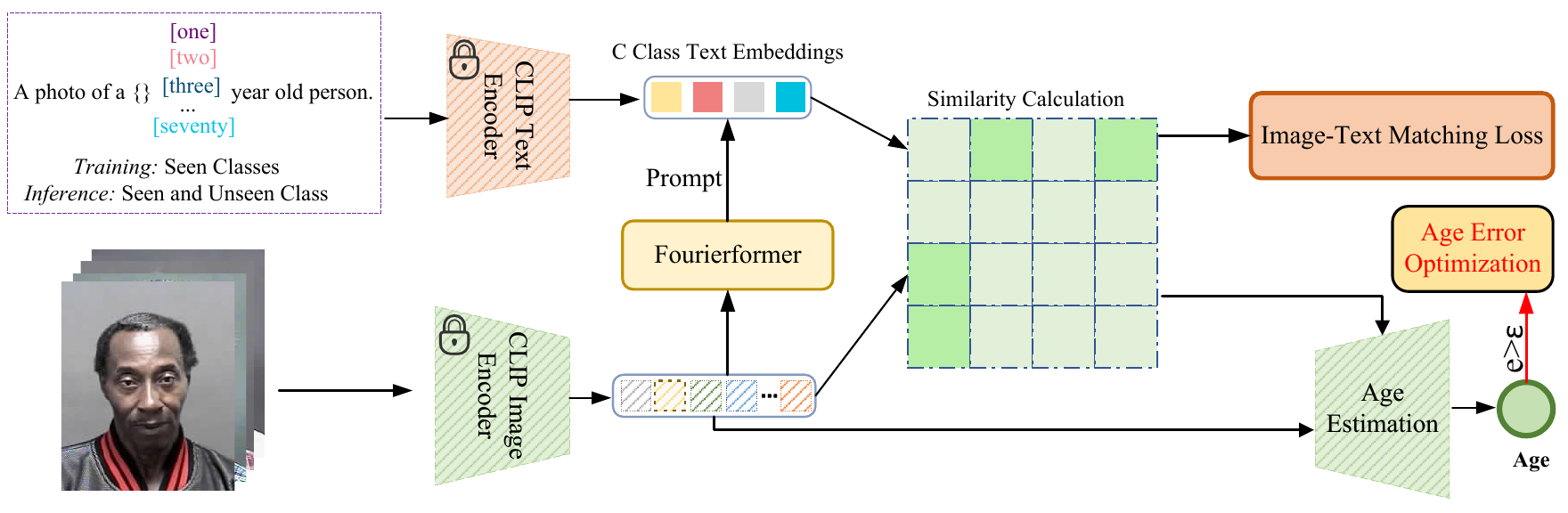}
	\caption{The overall framework for age prediction using CILF-CIAE. Specifically, we first use CLIP to extract image features and C-type text features, and then calculate the pixel-text similarity score. The similarity scores of the pixel-text pairs are fed into the age estimation module, and the age label is used as a supervision signal. To better utilize the prior knowledge of images, we introduce Fourierformer to extract contextual information in images to prompt the language model. Finally, we perform error optimization on the predicted age.}
	\label{fig:zero-shot}
\end{figure*}

\subsection{Contrastive Image-Language Pre-training}

With the powerful representation ability of the pre-trained visual-language model CLIP \cite{lee2022uniclip} in feature extraction learning, it has been widely used in CV tasks. CLIP uses a contrastive learning method to train by maximizing the similarity between the embedding vectors of related texts and images, which enables the model to find the most relevant text-image pairs in the embedding space and achieve natural language and image multimodal understanding. In the field of age estimation, we need a CLIP-based backbone network to directly perform inductive reasoning.

\section{METHODOLOGY}

\subsection{The Design of the CILF-CIAE Structure}
The CILF-CIAE architecture proposed in this paper is shown in Fig. \ref{fig:zero-shot}, which contains age prediction stages and age error optimization. Specifically, we first use age estimation models based CLIP with a Fourier prior module to predict the age of images. To further narrow the semantic gap between image and text features, we utilize an efficient contrastive multimodal learning module that supervises the multimodal fusion process of FourierFormer through contrastive loss for image-text matching, thereby improving the interaction effect between different modalities. Furthermore, if the predicted and actual values exceed a given threshold, the optimization branch is activated. The age errors are then used in the training of an ensemble error correction model to update the predicted age $x^\ast$. This training process continues until $e(x^\ast) \leq \epsilon$ terminates. The details of the CILF-CIAE architecture proposed in this paper will be described.

\subsubsection{Language-guided Visual Age Prediction}
As shown in Fig. \ref{fig:zero-shot}, we briefly introduce the CLIP-based visual language pre-training model for age estimation. CLIP consists of an image encoder and a text encoder \cite{zhang2022pointclip}. Image encoders aim to extract the underlying features of an image and map them into a low-dimensional embedding space. The architecture of image encoders usually uses ViT \cite{han2022survey} with superior performance. The text encoder often use Transformers \cite{khan2022transformers} to generate text representations with rich semantic information. Given a text prompt, such as ``A photo of a 12 year old person," the text encoder first converts each character into a lowercase byte-pair encoded representation, which uniquely identifies each character. The beginning and end of each text sequence are marked by [SOS] and [EOS]. Afterwards, the text representation is mapped into a 512-dimensional feature space, and then text Transformer is used for sequence modeling. Then, given an image feature obtained by the image encoder, the cosine similarity function is used to calculate the similarity between the image and the text prompt. The similarity formula is defined as follows:
\begin{gather}
	\mathbf{S}=\frac{\exp(sim(T_i,I_i)/\tau)}{\sum_{j=1}^N\exp(sim(T_j,I_i)/\tau)}
\end{gather}
where $\mathbf{S}$ is the similarity matrix, $T_i$ is the feature vector of the $i$-th text sequence obtained by the text encoder, $I_i$ is the feature vector of the $i$-th image obtained by the image encoder, $N$ represents the total number of training samples, $sim(\cdot)$ represents cosine similarity, and $\tau$ represents temperature attenuation coefficient.

To further narrow the semantic gap between image and text features, we design an efficient contrastive multimodal learning module that supervises the multimodal fusion process of FourierFormer through contrastive loss for image-text matching loss, thereby improving the interaction effect between different modalities. The optimization goal for image-text matching loss is defined as follows:
\begin{gather}
	\begin{aligned}
		\mathcal{L}_\text{text-image}& =-\frac1N\sum_{i=1}^N\log\frac{\exp(sim(T_i,I_i)/\tau)}{\sum_{j=1}^N\exp(sim(T_i,I_j)/\tau)} \\
		& + \frac1{N(N-1)}\sum_{i=1}^N\sum_{j\neq i}^N\log\frac{\exp(S(T_i,I_j)/\tau)}{\sum_{k=1}^N\exp(S(T_i,I_k)/\tau)}
	\end{aligned}
\end{gather}
where $N$ is the number of the training samples.

\subsubsection{Context-Aware Prompting}
Previous work has demonstrated that feature alignment of visual and language modalities can significantly improve the performance of CLIP models on downstream tasks \cite{zhang2022point}, \cite{zhou2022conditional}. Therefore, we consider whether we can design a customized context-aware prompting method to improve text features.

\textbf{Vision-to-language prompting.} The textual features that fuse visual global context information can make age estimation predictions more accurate. For example, ``a photo of a 68-year-old man with gray hair" is a more accurate prediction than ``a photo of a 68-year-old man." Therefore, we design a customized Fourier prior module to utilize visual global context information to improve text features in fine granularity. Specifically, we use the Fourierformer decoder to realize image spatial information interaction and channel evolution, and model the interaction between vision and language.

\begin{figure}
	\centering
	\includegraphics[width=0.5\linewidth]{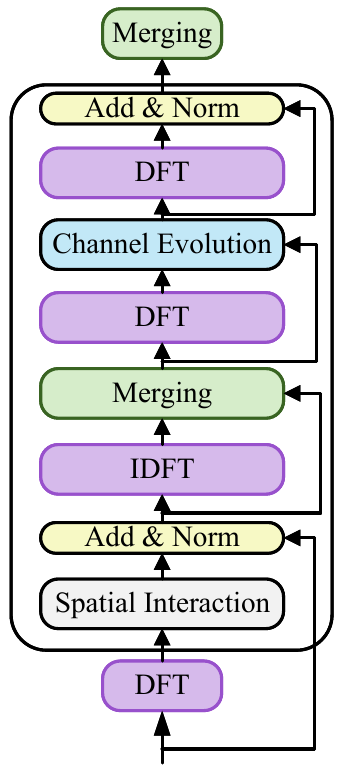}
	\caption{The overall framework of the proposed Fourierformer. FourierFormer includes a spatial interaction module, a channel evolution module, a discrete Fourier transform (DFT) and an inverse discrete Fourier (IDFT) module, which can effectively extract information from the global context of an image.}
	\label{fig:fourierformer}
\end{figure}

\subsubsection{Fourier Prior Embedded Block}
Fourier transform is used for frequency domain filtering, compression and feature extraction in image processing \cite{zhou2023fourmer}. By converting the image to the frequency domain, patterns and structures in the image can be more easily identified. For a given image $x \in \mathbb{R}^{H\times W\times C}$, the Fourier transform is applied to each image channel separately and transforms them into frequency domain space as complex components $\mathcal{F}(x)$. The formula is defined as:
\begin{gather}
	\mathcal{F}(\boldsymbol{x})(\boldsymbol{u},\boldsymbol{v})=\frac1{\sqrt{\mathrm{H\times W}}}\sum_{\boldsymbol{h}=0}^{\mathrm{H}-1}\sum_{\boldsymbol{w}=0}^{\mathrm{W}-1}\boldsymbol{x}(\boldsymbol{h},\boldsymbol{w})e^{-j2\pi(\frac h{\mathrm{H}}u+\frac{\boldsymbol{w}}{\mathrm{W}}\boldsymbol{v})}
\end{gather}
where $u$ and $v$ represent the horizontal and vertical coordinates of the Fourier domain. The phase component $P(x)(u,v)$ and the amplitude component $A(x)(u,v)$ are obtained as follows:
\begin{gather}
	\begin{aligned}
		&\mathcal{A}(\boldsymbol{x})(\boldsymbol{u},\boldsymbol{v})) =\sqrt{\mathcal{R}^2(\boldsymbol{x})(\boldsymbol{u},\boldsymbol{v}))+\mathcal{I}^2(\boldsymbol{x})(\boldsymbol{u},\boldsymbol{v}))},  \\
		&\mathcal{P}(\boldsymbol{x})(\boldsymbol{u},\boldsymbol{v})) =\arctan[\frac{\mathcal{I}(\boldsymbol{x})(\boldsymbol{u},\boldsymbol{v}))}{\mathcal{R}(\boldsymbol{x})(\boldsymbol{u},\boldsymbol{v}))}], 
	\end{aligned}
\end{gather}
where $I(x)(u, v)$ and $R(x)(u, v)$ represent imaginary numbers and real numbers, respectively.

\textbf{Structure Flow.} The main goal of designing the Fourier prior module in this paper is to achieve an effective and efficient global context image information modeling paradigm and improve the representation ability of text features, as shown in Fig. \ref{fig:fourierformer}. For a given image $x \in \mathbb{R}^{H \times W \times C_{in}}$, we first use a text encoder based CLIP to extract the shallow features of the image $X_0 \in \mathbb{R}^{H \times W \times C}$. Shallow features are encoded by using $N$ stacked image encoders. The Fuoriformer module designed in this paper consists of a stack of spatial interaction module, channel evolution module, residual and layer normalization module and Fourier prior module. Similarly, for the image decoder, we use a stack of the proposed core modules for image feature decoding.

\begin{figure*}
	\centering
	\includegraphics[width=1\linewidth]{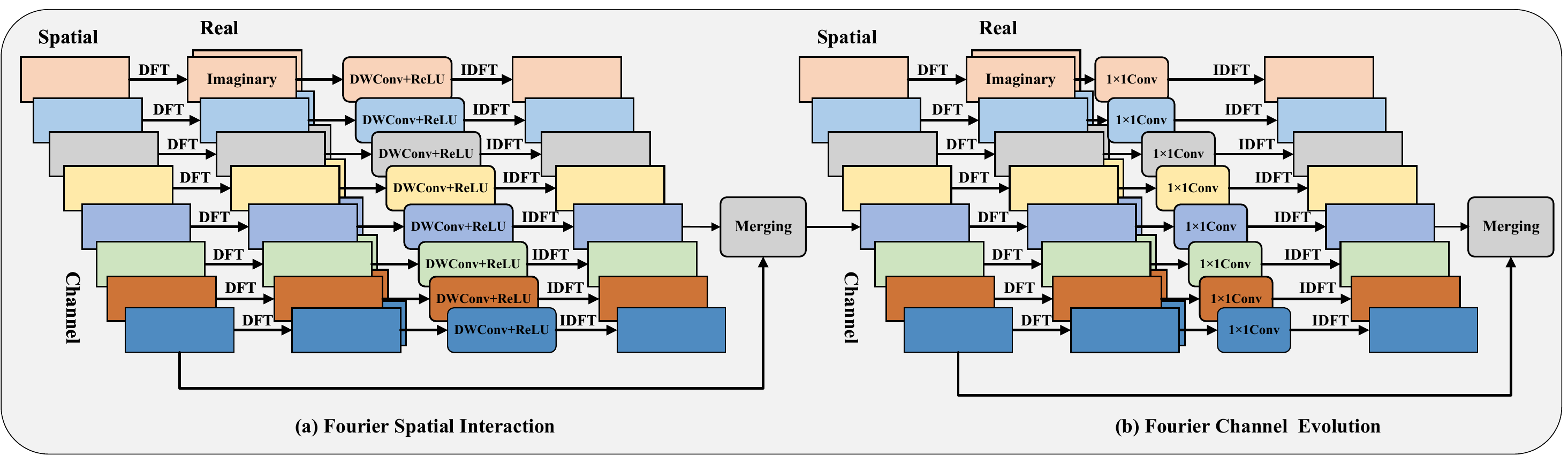}
	\caption{Details of the Fourier Prior Embedding module (FPE). FPE follows the global context information modeling idea of spatial interaction and channel evolution.}
	\label{fig:fourier}
\end{figure*}

As shown in Fig. \ref{fig:fourier}, the core module of Fourierformer consists of two parts: spatial interaction and channel evolution, which are implemented by depth convolution and $1\times 1$ convolution with DFT and IDFT respectively.

\textbf{Fourier Spatial Interaction.} Fourier spatial interaction first takes the image feature maps obtained by the image encoder as the input of Fourierformer, and then applies DFT to convert them into a spatial feature representation. Assuming that the features are expressed as $X \in R^{H\times W \times C}$, the corresponding DFT formula is defined as:
\begin{gather}
	\mathbf{X}_\mathcal{I}^{(\boldsymbol{c})},\mathbf{X}_\mathcal{R}^{(\boldsymbol{c})}=\mathcal{F}(\mathbf{X}^{(\boldsymbol{c})})
\end{gather}
where $c = 1, . . . , C$, $\mathbf{X}_\mathcal{I}$ and $\mathbf{X}_\mathcal{R}$ represent the real and imaginary parts in the Fourier space. We then perform Fourier spatial interaction to filter and compress the frequency domain signal of the image through a deep-wise convolution (DWconv) operation with LeakyReLU activation function. The spatial interaction process of images can be defined as:
\begin{gather}
	\begin{aligned}\mathbf{S}_{\mathcal{I}}^{(\boldsymbol{b})}&=LeakyReLU\left(DWconv^{(\boldsymbol{b})}(\mathbf{X}_{\mathcal{I}}^{(\boldsymbol{b})})\right)
		\\\mathbf{S}_{\mathcal{R}}^{(\boldsymbol{b})}&=LeakyReLU\left(DWconv^{(\boldsymbol{b})}(\mathbf{X}_{\mathcal{R}}^{(\boldsymbol{b})})\right)
	\end{aligned}
\end{gather}

Then we apply inverse DFT to the learned $\mathbf{S}_{\mathcal{I}}$ and $\mathbf{S}_{\mathcal{R}}$ with low-frequency signals to transform them back into the spatial domain. The formula for $\mathbf{S}_{\mathcal{I}}$ and $\mathbf{S}_{\mathcal{R}}$ to achieve time-frequency conversion is defined as follows:
\begin{gather}
	\mathbf{X_S^b}=\mathcal{F}^{-1}(\mathbf{S}_{\mathcal{I}}^{(\boldsymbol{b})},\mathbf{S}_{\mathcal{R}}^{(\boldsymbol{b})})
\end{gather}
The spectral convolution theorem in Fourier theory states that the convolution operation of signals in the frequency domain is equivalent to their product operation in the time domain, which reveals the overall frequency composition. The spectral convolution theorem provides an efficient way to process signals in the frequency domain because convolution operations in the frequency domain are generally easier to process than multiplication operations in the time domain. Therefore, we concatenate the $\mathbf{X_S^b}$ obtained by Fourier transform and normalize it to obtain the output $S_X$ of the Fourier spatial interaction.

\textbf{Fourier Channel Evolution.} Fourier channel evolution performs channel-by-channel evolution by applying a $1 \times 1$ convolution operator to decompose the output $S_X$ of the Fourier space interaction into real and imaginary parts $\mathbf{C}_{\mathcal{I}}$ and $\mathbf{C}_{\mathcal{R}}$. The Fourier channel evolution formula can be defined as:
\begin{gather}
	\begin{aligned}&\mathbf{C}\mathbf{X}_{\mathcal{I}}=LeakyReLU\left(\mathbf{con}\mathbf{v}\left(cat[\mathbf{C}_{\mathcal{I}}^1,\ldots,\mathbf{C}_{\mathcal{I}}^c]\right)\right)
		\\&\mathbf{C}\mathbf{X}_{\mathcal{R}}=LeakyReLU\left(\mathbf{con}\mathbf{v}\left(cat[\mathbf{C}_{\mathcal{R}}^1,\ldots,\mathbf{C}_{\mathcal{R}}^c]\right)\right)
	\end{aligned}
\end{gather}
where $cat(\cdot)$ is the concatenation operation. Then we perform IDFT to convert $\mathbf{C}\mathbf{X}_{\mathcal{R}}$ and $\mathbf{C}\mathbf{X}_{\mathcal{I}}$ to time domain space as follows:
\begin{gather}
	\mathbf{C_S^b}=\mathcal{F}^{-1}(\mathbf{C}\mathbf{X}_\mathcal{I}^{(\boldsymbol{b})},\mathbf{C}\mathbf{X}_\mathcal{R}^{(\boldsymbol{b})})
\end{gather}

\begin{figure*}
	\centering
	\includegraphics[width=1\linewidth]{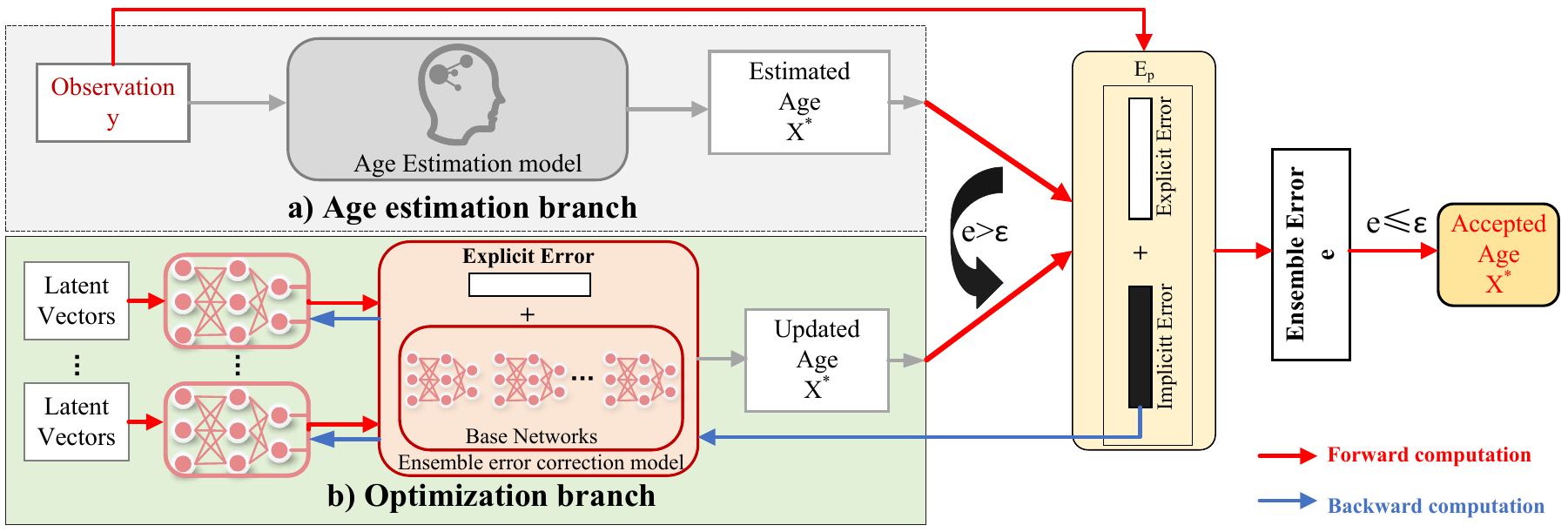}
	\caption{The flowchart of the correcting inverse age estimation. Existing age estimation models give a first age estimate, which is assessed by evaluations $E_P$. If failed, the optimization branch will be activated. The age estimation error estimated by the ensemble error model is used for training to update the predicted age $x^\ast$. The process terminates until $e(x^\ast) \leq \epsilon$.}
	\label{fig:error-correct}
\end{figure*}

\subsubsection{Two-stage Error Selection}
As shown in Fig. \ref{fig:error-correct}, we first use a CLIP-based learning model to predict age. If the error exceeds the threshold, an optimization branch is used to optimize the error and give a predicted age with high confidence.

For a given observation $y$, we use multiple models and metrics to evaluate the predicted age, resulting in an $h$-dimensional error vectors, expressed as:
\begin{gather}
	\mathbf{e}({\mathbf{x}},\mathbf{y})=\left[E_{1}({\mathbf{x}},\mathbf{y}),E_{2}({\mathbf{x}},\mathbf{y}),\ldots,E_{h}({\mathbf{x}},\mathbf{y})\right]
	\label{eq:9}
\end{gather}
where $E_i(,)$ represents the error estimate calculated by the $i$-th model, ${x}$ is the input image.

Each associated age estimate obtained from an observation $y$ follows the i.i.d. criterion, so $y$ is treated as a constant. Therefore, we can simplify Eq. \ref{eq:9} and obtain optimal model parameters by minimizing the error $e(x)$:
\begin{gather}
	\label{eq:10}
	\min_{{\mathbf{x}}\in\mathcal{X}}e({\mathbf{x}})=\sum_{i=1}^hw_iE_{i}({\mathbf{x}})
\end{gather}
where $w_i$ is determined using a voting mechanism, which is learnable.

Leveraging ensemble learning \cite{kang2023physics} enables a more robust representation of the hypothesis space, we integrate multiple neural networks to estimate implicit errors. Each neural network uses a mapping function $\phi(x, w)$, $\mathcal{R}^D\times{\mathcal{R}}^{|\mathbf{w}|}\to\mathcal{R}^k$ for error. We train $L$ regressors with the same network architecture and use a voting algorithm to obtain the final prediction. Therefore, for a given input state $x$, the implicit error $\hat{\mathbf{e}}$ is estimated by the ensemble network as follows:
\begin{gather}
	\hat{\mathbf{e}}\left({\mathbf{x}},\{\mathbf{w}_i\}_{i=1}^L\right)=\frac{1}{L}\sum_{i=1}^{L}\boldsymbol{\phi}\left({\mathbf{x}},\mathbf{w}_i\right)
	\label{eq:11}
\end{gather}
where $\mathbf{w}_i$ is the learnable network parameters.

According to Eq. \ref{eq:11}, we can obtain the cumulative age estimation error as follows:
\begin{gather}
	\begin{aligned}
		\hat{e}\left({\mathbf{x}},\{\mathbf{w}_i\}_{i=1}^L\right)&=\underbrace{\sum_{j=1}^kw_j\left(\frac1L\sum_{i=1}^L\boldsymbol{\phi}_j\left({\mathbf{x}},\mathbf{w}_i\right)\right)}_{\text{approximated implicit error}}
		\\ &+\underbrace{\sum_{j=k+1}^hw_jE_{j}({\mathbf{x}})}_{\text{true explicit error}}
	\end{aligned}
	\label{eq:12}
\end{gather}

We divide the error of Eq. \ref{eq:12} into two parts, one is the estimated implicit error, and the other is the true explicit error. The estimated implicit error is obtained by learning the feature representation of the image encoder by the ensemble regressor we built, and the real explicit error is obtained by the age estimation model based on CLIP we built. At the same time, we optimize the network parameters of the ensemble regressor by minimizing the distance between the estimated implicit error and the true explicit error. The optimization goal is defined as follows:
\begin{gather}
	\min_{\mathbf{w}_i}\mathbb{E}_{({\mathbf{x}},\mathbf{e})\sim D}\left[\mathrm{dist}\left(\boldsymbol{\phi}\left({\mathbf{x}},\mathbf{w}_i\right),\mathbf{e}_{1:k}\right)\right]
\end{gather}
where $\mathrm{dist}\left(\boldsymbol{\phi}\left({\mathbf{x}},\mathbf{w}_i\right),\mathbf{e}_{1:k}\right)=||\boldsymbol{\phi}\left({\mathbf{x}},\mathbf{w}_i\right),\mathbf{e}_{1:k}||^2_2$

To achieve controllable generation of predicted states, we use the feature representation decoded by the image encoder as the input of the ensemble regressor to learn and sample candidate predicted ages. Therefore, the update target of network parameters is defined as follows:
\begin{gather}
	\boldsymbol{\theta}^{(t)}=\arg\min_{\boldsymbol{\theta}\in\mathcal{R}^d}\mathbb{E}_{\mathbf{z}}\left[\hat{e}\left((\mathbf{z},\boldsymbol{\theta}),\left\{\mathbf{w}_i^{(t-1)}\right\}_{i=1}^L\right)\right]
\end{gather}
where $z$ is the the latent vectors. Finally, among the candidate age estimation states generated by the ensemble regressor with the trained network parameters $\theta^t$, we select the final prediction result as follows:
\begin{gather}
	{\mathbf{x}}_\mathrm{\Pi}^{(t)}=\arg\min_{{\mathbf{x}}\sim p\left({\mathbf{x}}|\boldsymbol{\theta}^{(t)}\right)}\hat{e}\left({\mathbf{x}},\left\{\mathbf{w}_i^{(t-1)}\right\}_{i=1}^{L}\right)
\end{gather}
The age estimation error is calculated via Eq. \ref{eq:9}. If the calculated error is less than the feasibility threshold, i.e. $\hat{e}(t) \leq \epsilon$, the selected age estimation state is considered acceptable and the predicted value is returned. Otherwise, the error is used to optimize the ensemble regressor model in the next iteration of parameter updates.

\subsection{Model Training}
Mean Absolute Error (MAE) is a commonly used performance evaluation metric in regression problems, which measures the mean absolute difference between model predictions and actual observations. The Loss is defined as follows:
\begin{gather}
	L^k(\theta)=|y^k-\hat{y}^k|
\end{gather}
where $\theta$ is the parameter of network learning, and $k$ represents the $k$-th training sample.

The optimization goals of the model are as follows:
\begin{gather}
	\min_{\theta}\sum_{k=1}^{N}L^k(\theta)
\end{gather}
Where $N$ represents the total number of samples.

\section{EXPERIMENTS}

\subsection{Benchmark Dataset Used}
In this paper, we use six benchmark datasets, MORPH-II\footnote{http://www.faceaginggroup.com/morph/}, FG-Net\footnote{http://yanweifu.github.io/FG\_NET\_data/FGNET.zip} CACD\footnote{http://bcsiriuschen.github.io/CARC/}, Adience\footnote{http://www.openu.ac.il/home/hassner/Adience/data.html}, FACES\footnote{http://faces.mpib-berlin.mpg.de}, and SC-FACE\footnote{https://www.scface.org/}, to conduct our age estimation experiments and verify the effectiveness of our CILF-CIAE method.

\textbf{MORPH-II.} The MORPH-II dataset is widely used in facial image research (e.g., age estimation and facial recognition). The MORPH-II dataset contains 55,000 facial photos of 13,000 volunteers over a period of time. The MORPH-II dataset covers facial images of volunteers from different ethnicities, different genders, and different geographical regions from 1 to 80 years old.

Existing methods employ three different experimental settings on the MORPH-II dataset. The first setting (S1) selects 5,492 white images from the original dataset (80\% images for training, 20\% images for testing) and performs 5-fold cross-validation to reduce cross-race effects \cite{rothe2018deep}, \cite{agustsson2017anchored}. The second setting (S2) randomly splits all images into training/test sets (80/20\%) and performs 5-fold cross-validation \cite{gao2017deep}. The third setting (S3) randomly selects 21,000 images from MORPH and restricts the black-white race ratio to 1:1 and the female to male ratio to 1:3 \cite{bao2023deep}.

\textbf{FGNET.} The FGNET dataset is composed of facial photos provided by volunteers from the age range of 0 to 69 years old. The FGNET dataset contains facial images of volunteers from different genders, different races, and different geographical areas. The FGNET dataset is mainly used to evaluate and improve the performance of facial age estimation algorithms.

\textbf{CACD.} CACD is also a dataset for facial age estimation, which mainly contains publicly available facial images of famous celebrities from social media (e.g., movies, TV, music). The CACD dataset contains more than 163,000 facial images of people from teenagers to older adults. The CACD dataset includes images of celebrities from different countries and different professions.

\textbf{Adience.} The Adience benchmark is an unconstrained dataset, i.e., there are no restrictions on gestures and photo poses. The face images in the Adience dataset are captured by mobile phone devices. Because these images are not subject to artificial data preprocessing and noisy image filtering, they can greatly reflect real-world challenges. The Adience dataset consists of 19,487 images, in which the numbers of males and females are 8,192 and 11,295 respectively.

\textbf{FACES.} The FACES face image dataset is a dataset used in psychology and neuroscience research, especially in studying age. This dataset was created by Ebner et al. in 2010 to provide a high-quality, diverse set of face images. The FACES dataset contains face photos of men and women ranging in age from 20 to 80 years old. The images show different emotional expressions such as happy, sad, angry and neutral expressions.

\textbf{SC-FACE.} SC-FACE (Surveillance Cameras Face Database) is a face image data set specially used for facial recognition research, especially facial recognition in surveillance environments. The dataset includes hundreds of images of subjects with facial expressions under different lighting conditions and backgrounds.

\subsection{Evaluation Metrics}
1) Mean Absolute Error (MAE): The MAE value reflects the absolute error between the true value of the sample and the predicted value of the model. In age estimation, MAE is more suitable as a model evaluation metric than MSE. The formula of MAE is defined as follows:
\begin{gather}
	MAE=\frac1N\sum_{i=1}^N|\hat{y}_i-y_i|
\end{gather}
where $\hat{y}$ represents the predicted value of the model, $y_i$ represents the true value, and $N$ represents the number of the samples.

2) Cumulative Score (CS): CS is used to measure the accuracy of the model's prediction error for face images not exceeding $L$ years. The formula for CS is defined as follows:
\begin{gather}
	CS(L)=(e_{\ell\leq L}/N)\times100\%
\end{gather}
where $e_{\ell\leq L}$ represents the number of samples where the absolute error $\ell$ of the model does not exceed $L$.

\subsection{Baseline Models}
\textbf{PML \cite{deng2021pml}:} Deng et al. proposed a progressive margin loss (PML) method to adaptively learn the distribution pattern of age labels. The PML method fully considers the inter-class and intra-class age distribution differences, and can effectively alleviate the long-tail distribution problem of data.

\textbf{Ranking-CNN \cite{chen2017deep}:} Chen et al. designed a novel Ranking-CNN architecture for age estimation. Ranking-CNN uses CNN to rank age labels and then perform high-level feature extraction. Ranking-CNN theoretically proves that the error comes from the maximum error in the ranked labels.

\textbf{DLDL \cite{gao2017deep}:} The deep label distribution learning (DLDL) method proposed by Gao et al. can adaptively learn the characteristics of label ambiguity. DLDL discretizes the age labels and uses CNN to minimize the KL divergence between the predicted distribution and the true distribution to optimize the model parameters.

\begin{figure*}[htbp]
	\centering
	\includegraphics[width=1\linewidth]{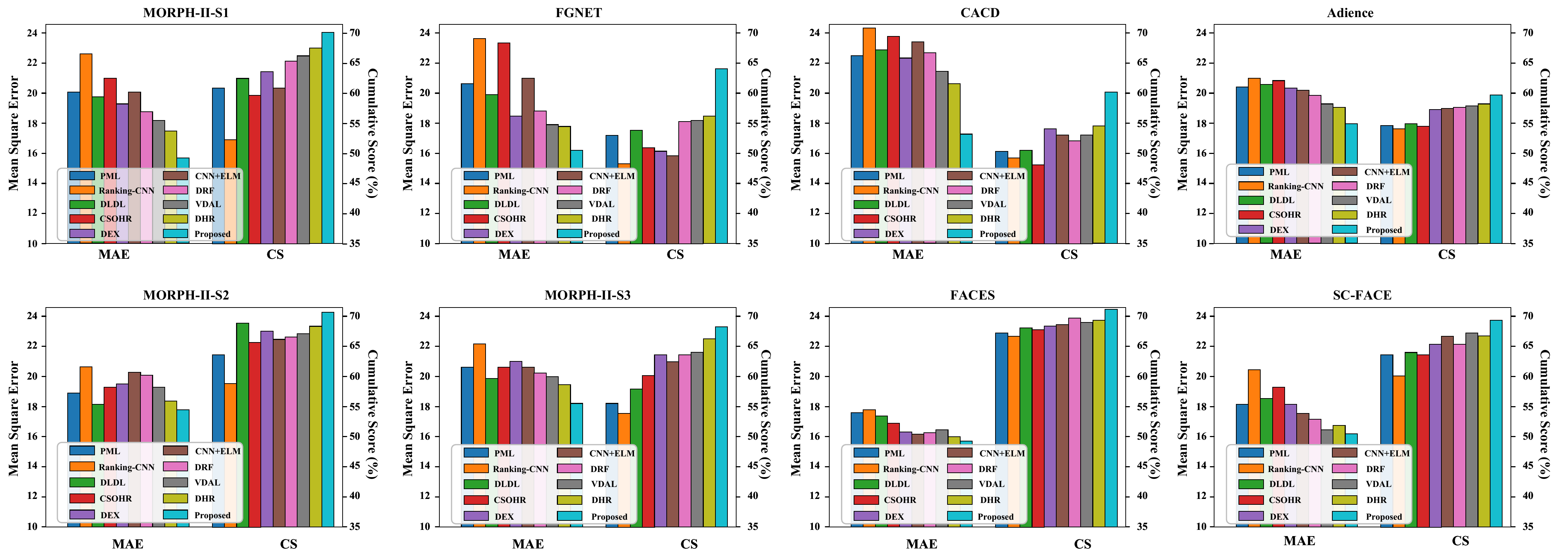}
	\caption{We tested the performance of our proposed method CILF-CIAE and some comparative methods on two evaluation metrics (i.e., MAE and CS) on six data sets and obtained corresponding experimental results.}
	\label{fig:data1}
\end{figure*}

\begin{figure*}[htbp]
	\centering
	\includegraphics[width=1\linewidth]{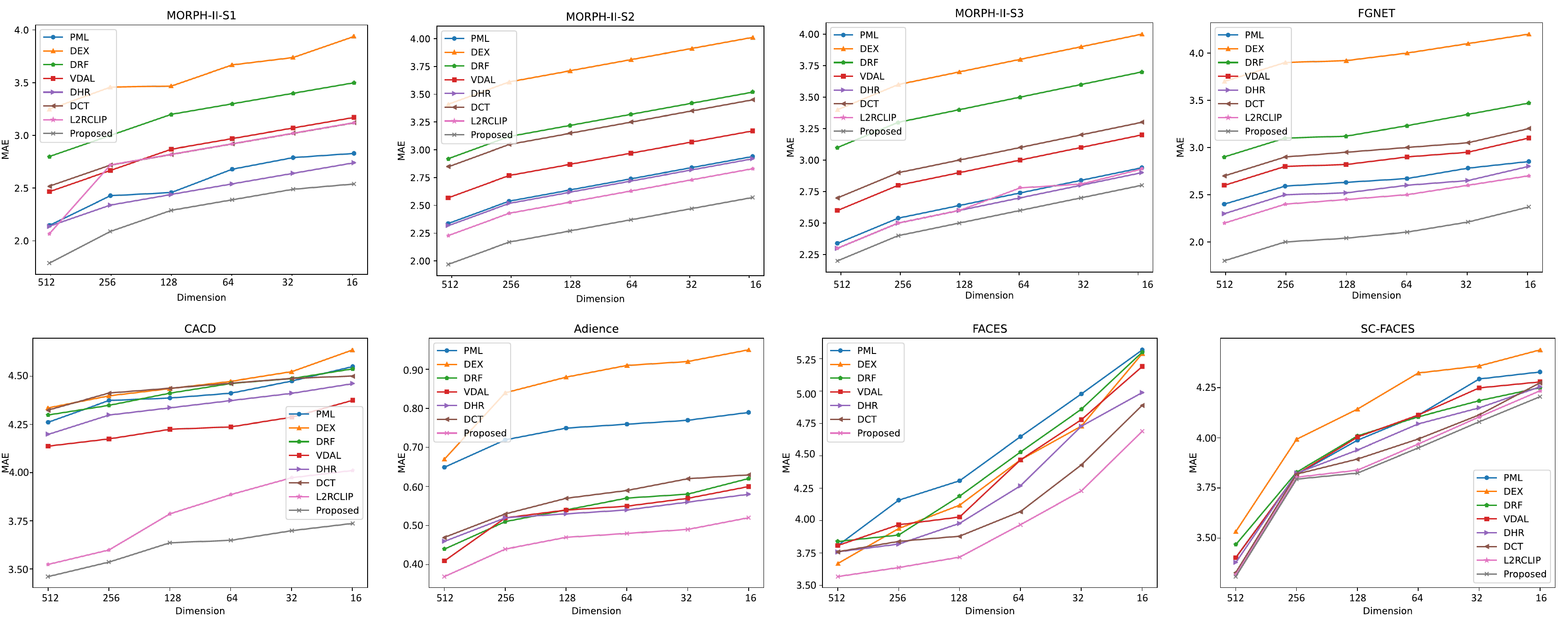}
	\caption{To explore the sensitivity of different models to parameters, we tested the impact of different feature embedding dimensions on CS on six data sets.}
	\label{fig:10}
\end{figure*}

\textbf{CSOHR \cite{chang2015learning}:} Chang et al. proposed a method combining hyperplane ranking algorithm and cost-sensitive loss for age estimation. CHOSR performs feature extraction on images with relative order information and introduces cost-sensitive losses to improve prediction accuracy.

\textbf{DEX \cite{rothe2018deep}:} The DEX proposed by Rothe et al. uses the VGG-16 architecture pre-trained on ImageNet for age estimation. DEX uses a deep CNN to align faces and age expectations to optimize model parameters.

\textbf{CNN+ELM \cite{duan2017ensemble}:} Duan et al. proposed a CNN and extreme learning machine (ELM) algorithm CNN2ELM for age estimation. CNN2ELM built three CNN networks to extract features and perform information fusion for Age, Gender and Race respectively, and then used ELM for the final age regression prediction.

\textbf{DRF \cite{shen2019deep}:} Shen et al. designed deep regression forest (DRF) for age estimation, which is continuously differentiable. DRF adaptively learns non-uniform age distribution data through the joint learning method of CNNC's random forest.

\textbf{VDAL \cite{liu2020similarity}:} Liu et al. proposed a similarity-aware deep adversarial learning (SADAL) method for age estimation. SADAL enhances the model's ability to learn facial age features through adversarial learning of positive and negative samples. In addition, SADAL designed a similarity-aware function to measure the distance between positive and negative samples to guide the optimization direction of the model.

\textbf{DHR \cite{tan2019deeply}:} Tan et al. proposed a deep hybrid alignment architecture for age estimation, which captures image age features with complementary semantic information through joint learning of global and local branches. Furthermore, in each branch network, a fusion mechanism is used to explore the correlation between sub-networks.

\textbf{DCT \cite{bao2022divergence}:} Bao et al. designed a divergence-driven consistency training mechanism to improve the quasi-efficiency of age estimation. DCT introduces an efficient sample selection strategy to select valid samples from unlabeled samples. Furthermore, DCT also introduces an identity consistency criterion to optimize the dependence between image features and age.

\begin{figure*}[htbp]
	\centering
	\includegraphics[width=1\linewidth]{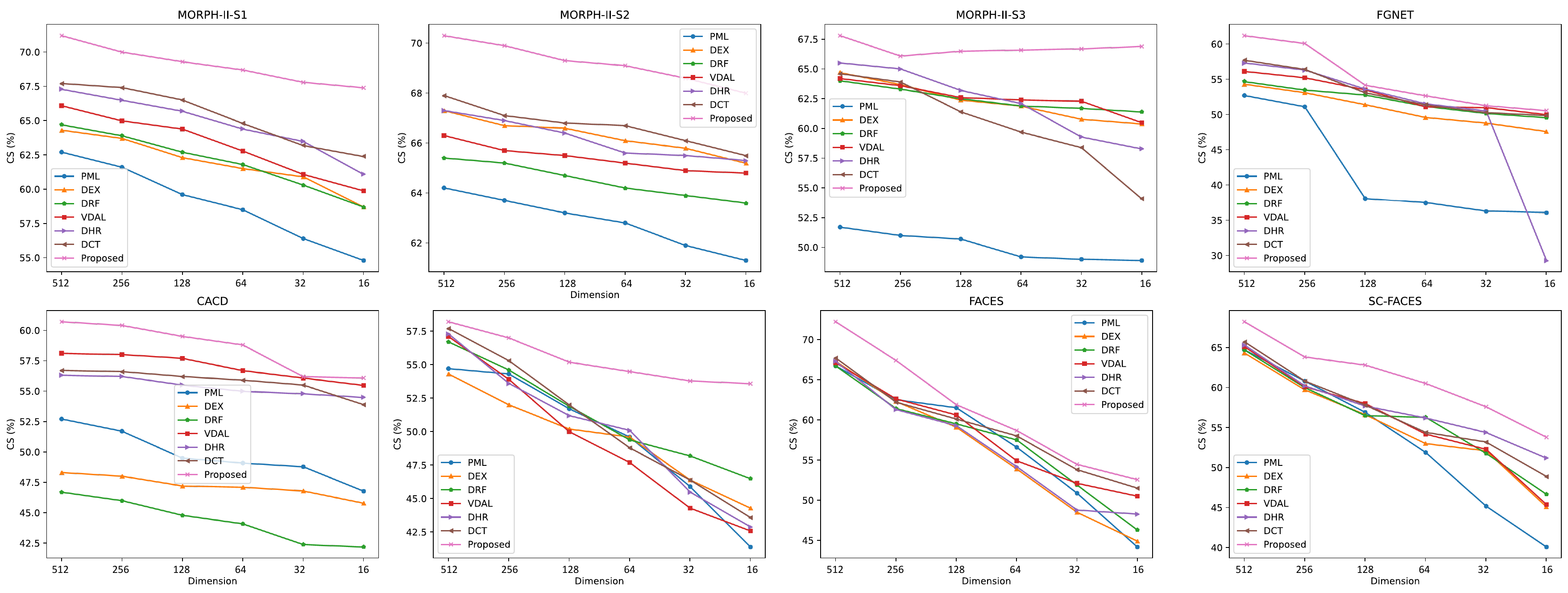}
	\caption{To explore the sensitivity of different models to parameters, we tested the impact of different feature embedding dimensions on CS on six data sets.}
	\label{fig:cs10}
\end{figure*}

\subsection{Implementation Details}
We adopt CLIP's pretrained image encoder as the backbone and directly integrate our designed Fourierformer as the decoder. In terms of language domain prompts, we choose a context length of 9. The Transformer decoder used to extract visual context consists of 6 layers. To reduce computational cost, we project image embeddings and text embeddings to 512 dimensions before the Transformer module. In terms of model fine-tuning, we observe that fine-tuning directly using the CLIP model does not produce satisfactory results. Therefore, we made a key modification: using AdamW as the optimizer for model training instead of the default SGD, which helps to improve the effectiveness of the training process and improve the final prediction performance.

\section{RESULTS AND DISCUSSION}
In this section, we discuss the experimental results of our method CILF-CIAE and other comparative methods on six data sets.

\subsection{Comparison with Baseline Methods}
To verify the superior performance of our proposed method CILF-CIAE, we conducted performance tests on six real data sets and compared it with other comparison methods. The experimental results are shown in Figs. \ref{fig:data1}. The method CILF-CIAE proposed in this paper has better MAE values and CS values on six data sets than other comparative methods. Specifically, the MAE values of CILF-CIAE under the three data set evaluation criteria of MORPH-S1, MORPH-S2 and MORPH-S3 are 1.74, 1.68 and 1.81 respectively, and the CS are 95.1\%, 95.7\% and 94.3\% respectively. Other comparison algorithms are worse than the CILF-CIAE algorithm in MAE value and CS value. Experimental results demonstrate that our method CILF-CIAE significantly outperforms other baseline algorithms. Similarly, on other data sets, our method CILF-CIAE method is also significantly better than other comparison algorithms. Experimental results show the robustness of the CILF-CIAE algorithm.

Overall, the feature learning ability of our method CILF-CIAE is better than other comparison algorithms in any case. Specifically, the performance improvement can be attributed to the high-quality text and image alignment capabilities based on the CLIP large model. Image representation based on language prompt guidance can greatly improve the ability to represent image features. At the same time, we introduce a context awareness module (i.e., Fourierformer) to react on language prompts to improve the expression of text semantic information. Unlike the traditional Vision Transformer architecture, Fourierformer models the global information of the image by introducing Fourier transform operations to achieve spatial interaction and channel evolution of image features. In addition, we also introduce an error correction mechanism. When the age predicted by the CLIP-based age estimation model differs greatly from the actual age, the model will start the optimization branch to optimize the error until $e(x)\leq\epsilon$ is reached.

\begin{table*}[htbp]
	\centering
	\caption{We perform ablation experiments to explore the impact of the three modules of spatial interaction, channel evolution, and error correction on age estimation performance respectively. We use six datasets to compare experimental results, and the MAE value is chosen as our evaluation metric.}
	\label{tab:ab1}
	\renewcommand\arraystretch{1.2}
	\scalebox{0.85}{
		\setlength{\tabcolsep}{2pt}{
			\begin{tabular}{ccccccccccc}
				\toprule
				Spatial interaction           & Channel evolution             & Error correction              & MORPH-S1 & MORPH-S2 & MORPH-S3 & FGNET & CACD & Adience & FACES & SC-FACES \\ \midrule
				\XSolidBrush    & \XSolidBrush    & \XSolidBrush    &  2.71        & 2.46         & 2.84         &  2.69     & 3.31    &  0.52       &  3.01     &  3.43         \\
				\CheckmarkBold & \XSolidBrush    & \XSolidBrush    & 2.63         &   2.31       &   2.69       &  2.62     &  3.24    & 0.47        &   2.86    & 3.35         \\
				\XSolidBrush    & \CheckmarkBold & \XSolidBrush & 2.65         &   2.34       &   2.69       &  2.67     &  3.26    & 0.48        &   2.83    & 3.36            \\
				\XSolidBrush    & \XSolidBrush    & \CheckmarkBold  & 2.44         &   2.17       &   2.48       &  2.41     &  3.13    & 0.44        &   2.67    & 3.14      \\
				\XSolidBrush    & \CheckmarkBold & \CheckmarkBold & 2.05         &   1.93       &   2.26       &  2.19     &  3.05    & 0.41        &   2.43    & 2.53       \\
				\CheckmarkBold & \XSolidBrush    & \CheckmarkBold & 1.91         &   1.85       &   2.14       &  2.06     &  2.94    & 0.39        &   2.38    & 2.40       \\
				\CheckmarkBold & \CheckmarkBold & \XSolidBrush   & 2.37         &   2.08       &   2.34       &  2.29     &  3.08    & 0.47        &   2.55    & 2.82       \\ 
				\rowcolor{gray!20} \CheckmarkBold & \CheckmarkBold & \CheckmarkBold &  \textbf{1.74}        &     \textbf{1.68}     &   \textbf{1.81}       &  \textbf{1.78}     &   \textbf{2.83}   &  \textbf{0.39}       &  \textbf{2.13}     &   \textbf{2.27}      \\ \bottomrule
	\end{tabular}}}
\end{table*}

\subsection{Effectiveness of Low-Dimensional Representation}
To explore the impact of the number of parameters of the model and the latent feature representation of the image on the model performance, we use different image feature dimensions (i.e., [512, 256, 128, 64, 32, 16]) to explore the effectiveness of low-dimensional representation. As shown in Figs. \ref{fig:10}, we tested the experimental effects of CILF-CIAE and other comparative methods on 6 data sets in different dimensions. We report the MAE values of the model. Specifically, the MAE value of CILF-CIAE increases slightly as the feature embedding dimension decreases on the six datasets, while the performance of other comparison methods drops sharply. Experimental results demonstrate the robustness of our method. The stable performance of CILF-CIAE may be attributed to the fact that the estimation algorithm based on CLIP contains rich image prior knowledge, which can improve the induction ability of the model. In addition, the Transformer architecture designed based on the Fourier change module to implement contextual prompts is a parameter-free estimation function and is insensitive to parameter changes.

As shown in Figs. \ref{fig:cs10}, we tested the experimental effects of CILF-CIAE and other comparative methods on six data sets in different dimensions. We report the CS values of the model. In tests on the MORPH-S1 and MORPH-S2 data sets, the CS value of CILF-CIAE decreased slightly as the image feature embedding dimension decreased. On other datasets, the CS value decreases rapidly with the decrease of image feature embedding dimension. However, the performance of CILF-CIAE is always higher than other comparison algorithms. The superior performance may be attributed to the optimization branch's ability to ensure that the prediction results are at a relatively high confidence level.

\begin{figure}
	\centering
	\includegraphics[width=1\linewidth]{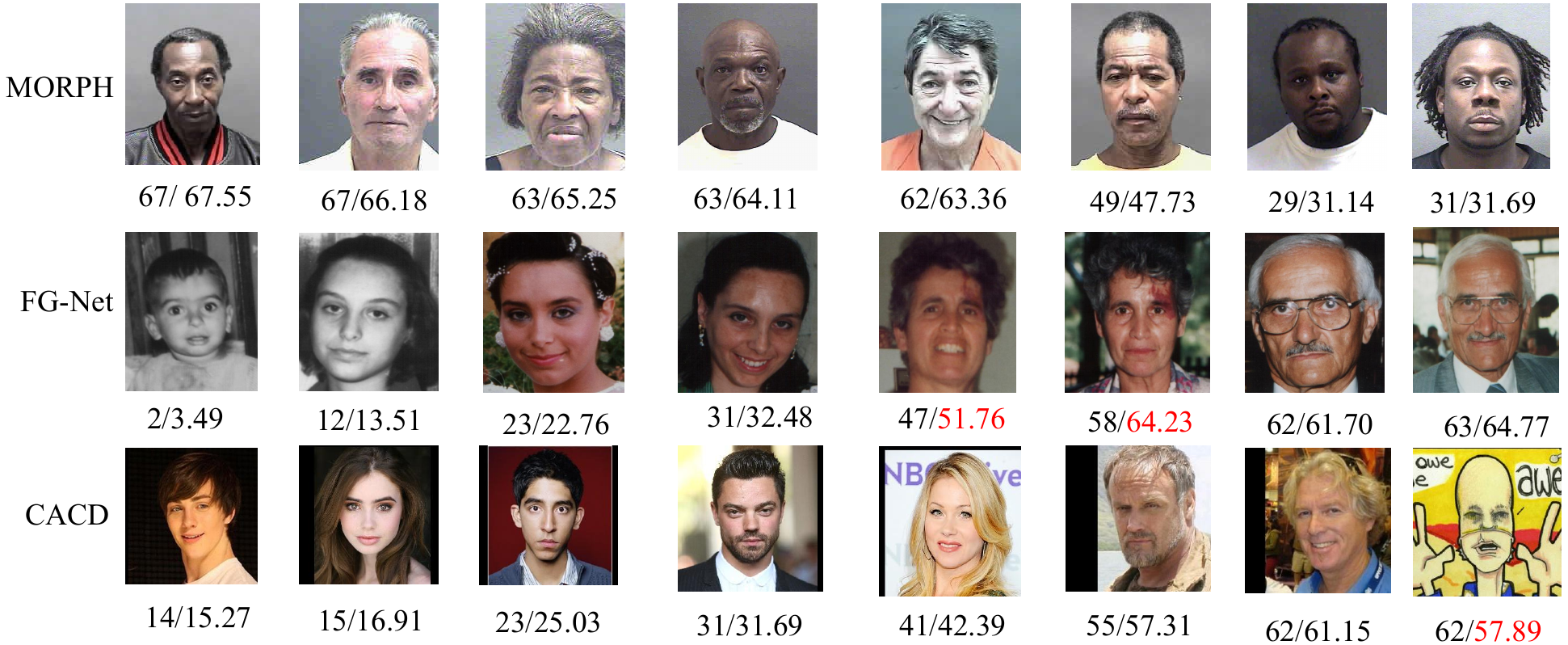}
	\caption{An example of age estimation results of our CILF-CIAE on the MORPH-II face dataset. The true labels are on the left and the estimated results are on the right. Poor estimation results are shown as red numbers.}
	\label{fig:agepredict}
\end{figure}

\subsection{Ablation Study}
As shown in Tables \ref{tab:ab1}, we perform ablation experiments on all test data respectively. We separately explored the effectiveness of the three modules proposed in this paper, i.e., spatial interaction module, channel evolution module and error correction module. If none of the three modules proposed in this paper are used, it means that the CLIP model is used directly to estimate the age of the image. The model has the worst experimental results on the six data sets if any of the modules proposed in this paper are not applied for age estimation. If one module is used for age estimation, the age estimation effect with the error estimation module is the best, the age estimation effect with the spatial interaction module is second, and the age estimation effect with the channel evolution module is the worst. When using two modules, the age estimation effect with the spatial interaction module and the error estimation module is the best, and the age estimation effect with the spatial interaction module and the channel evolution module is the worst. When three modules are used, the age estimation results are best in all cases. Ablation experiments demonstrate the effectiveness of each module proposed in this paper.

\begin{figure}
	\centering
	\includegraphics[width=1\linewidth]{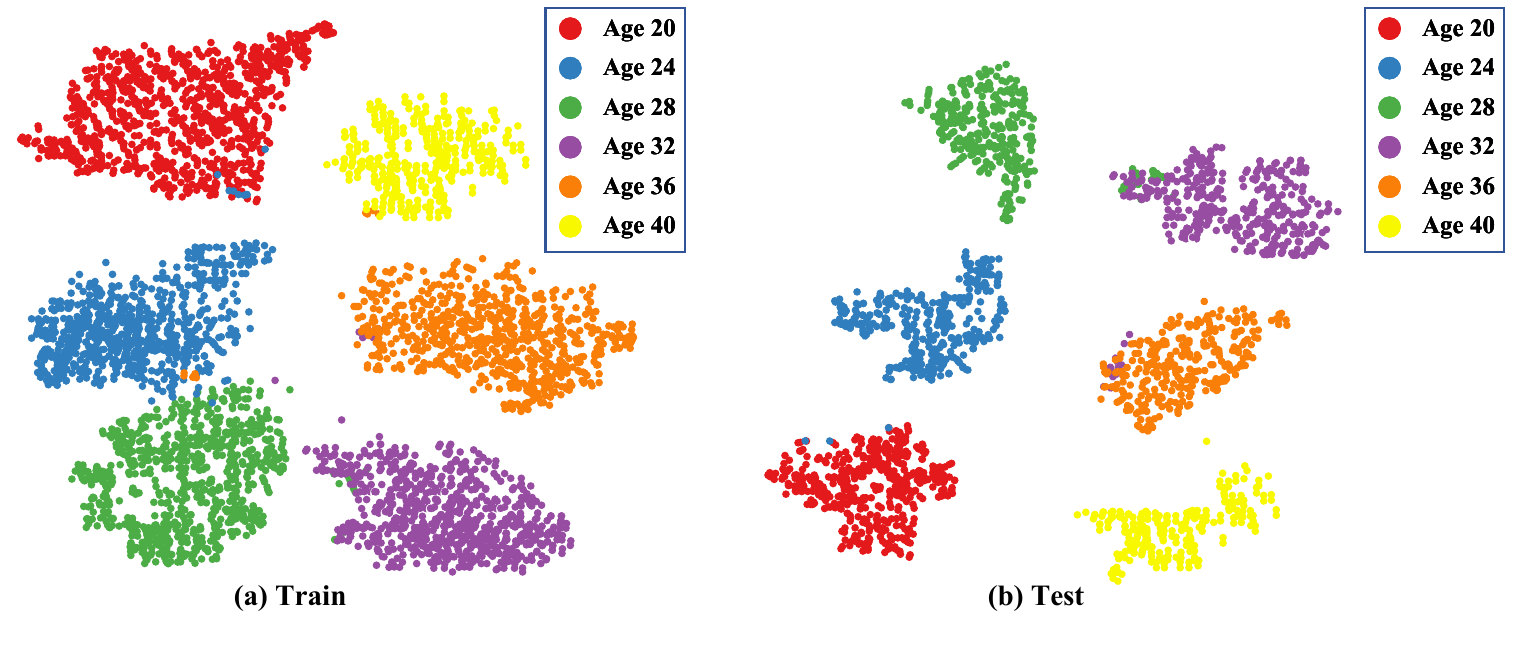}
	\caption{We visualize the learned features using t-SNE on the Morph II (S1) training and test sets. We visualize the distribution of the six age categories in the two-dimensional feature space.}
	\label{fig:agevisuf}
\end{figure}

\subsection{Qualitative Results Analysis}

To more intuitively demonstrate the effectiveness of CILF-CIAE, we conducted extensive experiments on the Morph-II benchmark dataset. Fig. \ref{fig:agepredict} shows the prediction results and true labels of CILF-CIAE. We observe that CILF-CIAE performs well in age prediction for most images and can accurately predict the age of faces. The inaccurate prediction on a few images may be attributed to the fact that the images are synthetic and the pose variations are large.

We further visualize the distribution of features learned in the training and testing phases on the Morph-II (S1) dataset using t-SNE. As can be seen from Fig. \ref{fig:agevisuf}, the feature class boundaries learned in the training and testing phases are relatively clear, and different age categories have more compact feature distributions.

\section{CONCLUSION AND FUTURE WORK}
The paper proposes a novel CLIP-driven Image–Language Fusion for Correcting  Inverse Age Estimation (CILF-CIAE) to perform age estimation. Firstly, we use Image Encoder and Text Encoder in CLIP to obtain corresponding feature representations and achieve age estimation. Secondly, we introduce a Transformer architecture based on Fourier transform to achieve spatial interaction and channel evolution of image features. Specifically, we replace the attention module in Transformer with Fourier transform and input image features into Fuorierformer to achieve spatial interaction and channel evolution. Finally, we build an error-correcting reversible age estimation module to ensure that the predicted age is within a high-confidence interval in an end-to-end learning manner. The method CILF-CIAE proposed in this paper achieves optimal age estimation on multiple age estimation datasets. In future research work, we will consider investigating estimation across data sets, which can improve the generalization ability of the model.

\section*{Acknowledgments}\label{sec8}
This work is supported by National Natural Science Foundation of China (Grant No. 69189338), Excellent Young Scholars of Hunan Province of China (Grant No. 20B625, No. 18B196), Changsha Natural Science Foundation (Grant No. kq2202294), and program of Research on Local Community Structure Detection Algorithms in Complex Networks (Grant No. 2020YJ009).


\bibliographystyle{elsarticle-num}
\bibliography{refs}


%
%
%

\end{sloppypar}
\end{document}